\documentclass{article}

\usepackage[numbers]{natbib}    
\usepackage{enumitem}
\usepackage{bbding}
\usepackage{pifont}
\usepackage[normalem]{ulem}
\usepackage[preprint]{neurips_data_2024}

\usepackage[utf8]{inputenc} 
\usepackage[T1]{fontenc}    
\usepackage{hyperref}       
\usepackage{url}            
\usepackage{booktabs}       
\usepackage{amsfonts}       
\usepackage{nicefrac}       
\usepackage{microtype}      
\usepackage{xcolor}         

\usepackage{graphicx} 
\usepackage{amsmath}

\usepackage{adjustbox}

\title{Hyperbolic Benchmarking Unveils Network Topology-Feature Relationship in GNN Performance}

\author{%
  Roya Aliakbarisani\thanks{These two authors contributed equally} \\
  Universitat de Barcelona \& UBICS \\
  \texttt{roya\_aliakbarisani@ub.edu}
  \And
  Robert Jankowski\footnotemark[1]  \\
  Universitat de Barcelona \& UBICS \\ 
  \texttt{robert.jankowski@ub.edu}  
  \And
   M. {\'A}ngeles Serrano \\
  Universitat de Barcelona, UBICS \& ICREA \\ 
  \texttt{marian.serrano@ub.edu}  
  \And
  Mari\'an Bogu\~{n}\'a\\
  Universitat de Barcelona \& UBICS \\ 
  \texttt{marian.boguna@ub.edu}
}

\begin{document}

\maketitle

\begin{abstract}
  Graph Neural Networks (GNNs) have excelled in predicting graph properties in various applications ranging from identifying trends in social networks to drug discovery and malware detection. With the abundance of new architectures and increased complexity, GNNs are becoming highly specialized when tested on a few well-known datasets. However, how the performance of GNNs depends on the topological and features properties of graphs is still an open question. In this work, we introduce a comprehensive benchmarking framework for graph machine learning, focusing on the performance of GNNs across varied network structures. Utilizing the geometric soft configuration model in hyperbolic space, we generate synthetic networks with realistic topological properties and node feature vectors. This approach enables us to assess the impact of network properties, such as topology-feature correlation, degree distributions, local density of triangles (or clustering), and homophily, on the effectiveness of different GNN architectures. Our results highlight the dependency of model performance on the interplay between network structure and node features, providing insights for model selection in various scenarios. This study contributes to the field by offering a versatile tool for evaluating GNNs, thereby assisting in developing and selecting suitable models based on specific data characteristics.
\end{abstract}

\section{Introduction}\label{sec: Intro}

Graph Neural Networks (GNNs)~\cite{scarselli2008graph, GNNBook2022, wu2021, zhou2020}, derived from Convolutional Neural Networks for graph-structured data, use recursive message passing between nodes and their neighbors. These models leverage graph topology and node-specific features to map nodes into a learnable embedding space.
GNNs have evolved to encompass a wide variety of architectures and tasks, ranging from node and graph classifications to link prediction. Despite this growing interest in the development of GNNs, the fundamental issue of homogeneity in benchmarking datasets persists in GNN research, making it challenging to determine the most suitable GNN model for unseen datasets. In addition, since GNNs are data-driven models tailored to specific tasks, there is a potential concern of overfitting new architectures to given datasets, especially when the data have similar structural properties~\cite{Palowitch2022GraphWorld}. Thus, a fair comparison between different models in reproducible settings is required.

In this work, we propose a comprehensive benchmarking scheme for graph neural networks. Utilizing a \textbf{Hyp}erbolic Soft Configuration \textbf{N}etwork Model with \textbf{F}eatures (HypNF) \cite{aliakbarisani2023featureenriched}, we generate synthetic networks with realistic topological properties where node features can be correlated with the network topology. This highly flexible model allows us to evaluate GNNs in depth across various scenarios. Moreover, we suggest using the benchmark as a tool for optimal model selection by analyzing the inherent properties of the real dataset. Although the use of hyperbolic geometry might appear superfluous, it has been demonstrated that it is the simplest method for generating geometric random graphs that uniquely combine several key characteristics: They have power law degree distributions, exhibit small-world properties, and are highly clustered, meaning they have a high density of triangles~\cite{boguna2021network}. These traits closely mirror those observed in real complex networks.

We aim to show the crucial factors, including the network's structural properties and the degrees of correlation between nodes and features—controlled by the framework parameters—that influence the performance of graph machine learning models. Employing the proposed benchmark, we systematically compare the performance of well-known GNNs and evaluate models that solely utilize node features. Our study aims to evaluate machine learning models in two fundamental graph-based tasks: node classification (NC) and link prediction (LP).

Here, we highlight the main contributions of our empirical study, which provides insights into the suitability of the various models under different network conditions. It will, thus, benefit applications and the community focused on developing new GNN algorithms.
\begin{itemize}[leftmargin=*]
    \item Our framework generates benchmark networks with tunable levels of topology-feature correlation, homophily, clustering, degree distributions, and average degrees. This approach covers the most important properties of a wide range of real datasets, providing a comprehensive tool for their analysis. The code and the datasets will be publicly available at \url{https://github.com/networkgeometry/hyperbolic-benchmark-gnn} under the MIT License.
 
    \item GNNs exhibit varying levels of performance fluctuation under a fixed set of parameters. Notably, HGCN~\cite{Chami2019Hyperbolic}
shows less robustness compared to GCN~\cite{Kipf2017Semi} when the network's average degree is low. However, this trend reverses in networks with homogeneous degree distributions.
    
    \item The stronger the correlation between the network topology and node features, the better GNNs and feature-based models perform in NC and LP.
        
    \item The hyperbolic-based models, specifically HGCN and HNN~\cite{ganea2018hyperbolic}, achieve the highest AUC scores~\cite{Melo2013} in LP task. Remarkably, HNN, despite being solely a feature-based method, outperforms traditional graph-based models across various parameters.
 
    \item In the NC task, where no information about the graph data is available, emphasis should be placed on model interpretability and time complexity. This is crucial since the accuracy of graph-based models tends to be uniformly high, making these factors significant differentiators.
\end{itemize}

\section{Related work} \label{sec: Related work}

With the continuous evolution of graph machine learning, there is a growing necessity to comprehend and evaluate the performance of GNN architectures. In this respect, benchmarking can provide a fair and standardized way to compare different models. The Open Graph Benchmark (OGB)~\cite{hu2020open} stands as a versatile tool to assess the performance of the GNNs. Yet, its emphasis on a limited range of actual networks indicates that it does not encompass all network characteristics and falls short in terms of parameter manipulation. Consequently, this highlights the necessity for creating benchmarking tools based on synthetic data. Such tools would allow for the assessment of GNNs in a controlled environment and across a more extensive array of network properties~\cite{Wang2021FastSGG, MAEKAWA2023102195,maekawa2022beyond}. One of them is GraphWorld~\cite{Palowitch2022GraphWorld}, which is a synthetic network generator utilizing the Stochastic Block Model (SBM) to generate graphs with communities. It employs a parametrized community distribution and an edge probability matrix to randomly assign nodes to clusters and establish connections. Node features are also generated using within-cluster multivariate normal distributions. A fixed edge probability matrix in SBM prevents GraphWorld from faithfully replicating a predefined degree sequence and generating graphs with true power-law distributions. To overcome this limitation, Ref.~\cite{yasir2023examining} integrates Graphworld with two other generators: Lancichinetti-Fortunato-Radicchi (LFR)~\cite{Lancichinetti2008Benchmark} and CABAM (Class-Assortative graphs via the Barab\'asi-Albert Model)~\cite{cabam2020shah}. This integration broadens the coverage of the graph space, specifically for the NC task. In this paper, we propose an alternative  network generators: a framework based on the geometric soft configuration model. This model's underlying geometry straightforwardly couples the network topology with node features and labels. This capability enables independent control over the clustering coefficient in both the unipartite network of nodes and the bipartite network of nodes and features, irrespective of the degree distributions of nodes and features (see Fig.~\ref{fig:Param_independency} in Appendix~\ref{APX:Degree&Clustering control}). Table \ref{tab:comparison} presents a comparison between HypNF and several state-of-the-art benchmarking frameworks, highlighting the properties each can control.

\begin{table}[t]
	\centering
	\caption{Comparison of HypNF and other synthetic network generators. The \checkmark indicates control over a given property, \ding{55} indicates lack of control, and $\propto$ indicates indirect control. The $P(k)$ is the degree distribution, $\langle c \rangle$ is the clustering coefficient, $\langle k \rangle$ is the average degree. Whereas $P(k_n)$ and $P(k_f)$ are the degree distributions of nodes and features, respectively, $\langle k_n \rangle$ and $\langle k_f \rangle$ are the average degrees of nodes and features, respectively, and $\mathcal{H}$ represents homophily.}
	\label{tab:comparison}
	\begin{adjustbox}{width=1\textwidth}
		\begin{tabular}{@{}lcccccccc@{}}
			\toprule
			& $P(k)$, Power-law &  $\langle c \rangle$ & $\langle k \rangle$ & $P(k_n),P(k_f)$ & $\langle k_n \rangle, \langle k_f \rangle$ & \begin{tabular}[c]{@{}c@{}}Topology-Feature\\ Correlation\end{tabular}  & $\mathcal{H}$ & \begin{tabular}[c]{@{}c@{}}Underlaying mechanism\slash  \\  Connectivity law\end{tabular} \\ \midrule
			HypNF  & \checkmark, \checkmark & \checkmark  & \checkmark & \checkmark,\checkmark & \checkmark,\checkmark & \checkmark & \checkmark & geometric \\ 
			GraphWorld (SBM)~\cite{Palowitch2022GraphWorld}     & \checkmark, Quasi power-law & $\propto$ & \checkmark  & \ding{55}, normal & \ding{55}, \checkmark  & $\propto$ & \checkmark &  probability matrix \\
			GraphWorld (CABAM)~\cite{yasir2023examining}  & \ding{55}, \checkmark  & $\propto$  & \checkmark  & \ding{55}, normal & \ding{55}, \checkmark  & $\propto$ & \checkmark & preferential attachement \\ 
			GraphWorld (LFR)~\cite{yasir2023examining}  & \checkmark,\checkmark  & $\propto$  & \checkmark  & \ding{55}, normal & \ding{55}, \checkmark  & $\propto$ & \checkmark & class-aware configuration model \\ 	
			GenCAT~\cite{MAEKAWA2023102195,maekawa2022beyond}   & \checkmark, \checkmark & $\propto$  & \checkmark & \ding{55}, (normal, Bernoulli) & \ding{55}, \checkmark & $\propto$ & \checkmark & latent variables \\
			FastSGG~\cite{Wang2021FastSGG}       & \checkmark, \checkmark &  $\propto$ & \checkmark &  \ding{55} , \ding{55}&  \ding{55}, \ding{55}&  \ding{55} & \checkmark & preferential attachment \\ \bottomrule
		\end{tabular}
	\end{adjustbox}
\end{table}

\section{HypNF Model} \label{sec:Model}
Graph-structured data are characterized by group of $N_n$ nodes that create a complex network $\mathcal{G}_n$, along with a collection of $N_f$ features linked to these nodes. Typically, these features are converted into binary form. Therefore, for any given node $i$, its feature set is depicted as a vector $\vec{f}_i \in \{0, 1\}^{N_f}$. This vector has elements that are either zero or one, depending on whether a specific feature is present or absent. An alternative approach considers that nodes and features form a bipartite graph, $\mathcal{G}_{n,f}$, with nodes defining one of the types of elements of the graph and features the other~\cite{aliakbarisani2023featureenriched}. Within this approach, the full information of the data is encoded into the two networks $\mathcal{G}_n$ and $\mathcal{G}_{n,f}$. Interestingly, in~\cite{aliakbarisani2023featureenriched}, the topological properties of the bipartite graph $\mathcal{G}_{n,f}$ of real graph-structured datasets were first studied, showing a rich and complex topological organization. 

A remarkable finding in~\cite{aliakbarisani2023featureenriched, jankowski2024feature} is the detection of strong correlations between the graphs $\mathcal{G}_n$ and $\mathcal{G}_{n,f}$ within real datasets. This suggested the possibility to describe $\mathcal{G}_{n,f}$ as a geometric random graph in the same hyperbolic space used to describe the graph $\mathcal{G}_n$ so that the shared metric space would mediate the correlation between them. Building on this concept, the study in~\cite{aliakbarisani2023featureenriched} introduces a generative model that produces networks $\mathcal{G}_n$ and $\mathcal{G}_{n,f}$, with experimental results demonstrating their ability to accurately reproduce key topological properties observed in real datasets, including degree distribution, clustering coefficient, and average nearest neighbor degree function (see Fig.~\ref{fig:Cora_Properties} in Appenix~\ref{APX:Empirical validation}). We will now provide a detailed description of this model, which is used to create synthetic datasets for evaluating the performance of GNNs.

\begin{figure*}[t]
	\centering
	\includegraphics[width=0.9\linewidth]{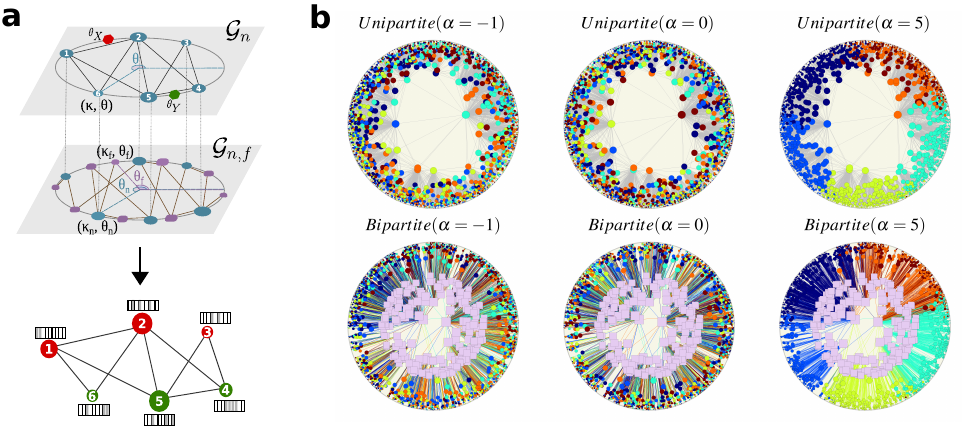}
	\caption{(a) Representation of HypNF benchmarking framework to generate rich graph-structured data. (b) Hyperbolic representation of a synthetic network with $N_n = 2000$ nodes represented as circles, where colors indicate their labels, and $N_f = 200$ features depicted as purple squares. The size of the symbols is proportional to the logarithms of the degrees of nodes and features. The parameters for the $\mathbb{S}^1$ model are $\beta = 3$, $\gamma = 3.5$, and $\langle k \rangle = 30$, and for the bipartite-$\mathbb{S}^1$ model, $\beta_b = 3$, $\gamma_n = 3.5$, $\gamma_f = 2.1$, and $\langle k_n \rangle = 3$. Only edges with an effective distance $\chi < 1$ are depicted in the figure.}	
	\label{fig:1}
\end{figure*}
\subsection{The $\mathbb{S}^1/\mathbb{H}^2$ model}

The network $\mathcal{G}_n$ is modeled by the $\mathbb{S}^1/\mathbb{H}^2$ model~\cite{serrano2008similarity,krioukov2010hyperbolic,serrano_boguna_2022}. This geometric soft configuration model produces synthetic networks with realistic topological properties, including arbitrary degree distributions~\cite{serrano2008similarity,krioukov2010hyperbolic,gugelmann2012random}, the small-world property~\cite{abdullah2017typical,friedrich2018diameter,muller2019diameter}, self-similarity~\cite{serrano2008similarity}, and high clustering~\cite{Boguna:2020fj,krioukov2010hyperbolic,gugelmann2012random,candellero2016clustering,Fountoulakis2021}, to name just a few. See also~\cite{boguna2021network} and references therein.

In the $\mathbb{S}^1$ model a node is endowed with a hidden degree $\kappa$, representing its popularity, influence or importance. Hidden (or expected) degrees are distributed by an arbitrary probability density $\rho(\kappa)$, with $\kappa \in (\kappa_0,\infty)$. In this way, the model has the flexibility to reproduce a variety of degree distributions. Each node is also assigned with an angular position $\theta$ in the similarity space, represented as a one-dimensional sphere~\footnote{The model with arbitrary dimensions has been defined in~\cite{jankowski2023d}.}. Then, pairs of nodes are connected with a probability function taking the form of a gravity law, balancing the interplay between node angular distances and their hidden degrees:
\begin{equation}
    p(\kappa,\kappa', \Delta \theta)=\frac{1}{1+\chi^\beta} \;  \; \mbox{with}  \; \; \chi \equiv \frac{R \Delta \theta}{\mu \kappa \kappa'},
    \label{Eq:S1probCon}
\end{equation}
where $\Delta \theta = \pi - |\pi - |\theta - \theta'||$ is the angular separation between the nodes in the similarity space, $R =  N_n/2\pi$ denotes the radius of the similarity space, $\beta > 1$ governs the level clustering, and 
\begin{equation}
\mu = \frac{\beta}{2\pi \langle k \rangle} \sin{\frac{\pi}{\beta}} 
\label{eq:mu}
\end{equation}
controls the average degree of the network $\langle k \rangle$. Notice that the radius is chosen such that in the limit $N_n \gg 1$ the curvature of the similarity space vanishes and the process converges to a Poisson point process on $\mathbb{R}$ of density one. With these choices, the expected degree of a node with hidden degree $\kappa$ is $\bar{k}(\kappa)=\kappa$, so that by controlling the distribution of hidden degrees, we can adjust the resulting degree distribution.

Interestingly, the $\mathbb{S}^1$ model exhibit an isomorphism with a purely geometric model in hyperbolic space, referred to as the $\mathbb{H}^2$ model. The isomorphism is realized by mapping nodes' hidden degrees into radial positions within a disk of radius $R_{\mathbb{H}^2}$ in the hyperbolic plane as
\begin{equation}
r = R_{\mathbb{H}^2}-2\ln{\frac{\kappa}{\kappa_0}},~~~~\kappa \ge \kappa_0,
\end{equation}
where the radius in $\mathbb{H}^2$ is given by $R_{\mathbb{H}^2} = 2\ln{\frac{2R}{\mu \kappa_0^2}}$. Low degree nodes with hidden  degree $\kappa_0$ are mapped at the boundary of the hyperbolic disk, whereas high degree nodes get located close to the center  of  the disk. After this mapping, the connection probability in Eq.~\eqref{Eq:S1probCon} becomes 
\begin{equation}
p(x)=\frac{1}{1+e^{\frac{\beta}{2}(x-R_{\mathbb{H}^2})}} \; \; \mbox{, with} \; \; x=r+r'+2\ln{\frac{\Delta \theta}{2}} 
\label{eq:pkkhyperbolic}
\end{equation}
where $x$ approximates the hyperbolic distance between two nodes at radial coordinates $r$ and $r'$ with angular separation of $\Delta \theta$. Thus, in this representation, the probability of connections is just a function of the hyperbolic distance.

\subsection{The bipartite-$\mathbb{S}^1/\mathbb{H}^2$ model}

In real-world graph-structured data, nodes' features are correlated with the topology of the graph, i.e., nodes that exhibit similarities in the network topology $\mathcal{G}_n$ also share common features~\cite{ali2021predicting,aliakbarisani2023featureenriched}. The bipartite-$\mathbb{S}^1$ model induces this correlation by placing both $\mathcal{G}_n$ and $\mathcal{G}_{n,f}$ in the same similarity space. 

In this model, every node is given a pair of hidden variables, $(\kappa_n, \theta_{n})$. Here, $\kappa_n$ denotes the node's expected degree in the bipartite node-feature graph, while $\theta_n$ (equal to $\theta$) represents its angular coordinate, matching that in $\mathcal{G}_n$. In a similar fashion, features are assigned two hidden variables, $(\kappa_f, \theta_f)$, accounting for their expected degrees and their angular locations in a the shared similarity space. The likelihood of a link forming between a node and a feature, characterized by hidden degrees $\kappa_n$ and $\kappa_f$ and separated by an angular distance $\Delta \theta$, is expressed as
\begin{equation}
p_b(\kappa_n,\kappa_f,\Delta \theta)=\frac{1}{1+\chi^{\beta_b}}, \; \; \text{where} \; \; \chi \equiv \frac{R \Delta \theta}{\mu_b \kappa_n \kappa_f}.
\end{equation}
In this equation, 
\begin{equation}
\mu_b=\frac{\beta_b}{2\pi \langle k_n \rangle} \sin{\frac{\pi}{\beta_b}}
\label{eq:mub}
\end{equation}
is a crucial parameter that defines the average degree of nodes $\langle k_n \rangle$ and features $\langle k_f \rangle=\frac{N_n}{N_f}\langle k_n \rangle$ and $\beta_b$ controls bipartite-clustering, as a measure of the coupling between the resulting topology and the underlying metric space. This model, akin to the $\mathbb{S}^1$ model, guarantees that the expected degrees of nodes and features with hidden degrees $\kappa_n$ and $\kappa_f$ are $\bar{k}_n(\kappa_n)=\kappa_n$ and $\bar{k}_f(\kappa_f)=\kappa_f$, respectively. The hidden variables for both nodes and features can be generated from arbitrary distributions or tailored to mimic the topology of a specific real-world network.

As in the case of the $\mathbb{S}^1$ model, nodes and features in $\mathcal{G}_{n,f}$ can be mapped into the hyperbolic disc of radius $R^b_{\mathbb{H}^2}=2\ln{\frac{2R}{\mu_b \kappa_{n,0}\kappa_{f,0}}}$ through the following transformations

\begin{equation}
r_n=R^b_{\mathbb{H}^2}-2\ln{\frac{\kappa_n}{\kappa_{n,0}}}\;\; ~~\mbox{and}~~ \; \;r_f=R^b_{\mathbb{H}^2}-2\ln{\frac{\kappa_f}{\kappa_{f,0}}},
\end{equation}
where $\kappa_{n} \ge \kappa_{n,0}$ and $\kappa_{f} \ge \kappa_{f,0}$. This mapping leads to a connection probability between nodes and features with the same functional form as in Eq.~\eqref{eq:pkkhyperbolic}, replacing $\beta$ by $\beta_b$ and $R_{\mathbb{H}^2}$ by $R^b_{\mathbb{H}^2}$.

\subsection{Assigning labels to nodes}

To complete our theoretical framework, we need to specify how labels are assigned to nodes, ensuring a correlation between node labels in $\mathcal{G}_n$ and their features. We use the underlying similarity space to induce and tune these correlations. We introduce a node labeling strategy for any number of clusters $\mathcal{N}_L$, each associated with a label, and a tunable homophily level~\cite{zhu2020beyond,Pei2020Geom-GCN,lim2021new}, crucial for GNNs~\cite{ma2021homophily,wang2024understanding}. First, the cluster centroids are randomly placed in the similarity space. The probability of node $i$ being assigned to cluster $X$ is given by:
\begin{equation}
	\label{eq:1}
	P_X(i) = \frac{\Delta\theta_{iX}^{-\alpha}}{\sum\limits_{Y\in \mathcal{N}_L} \Delta\theta_{iY}^{-\alpha}}.
\end{equation}
where $\Delta\theta_{iX}$ is the angular distance between node $i$ and centroid $X$. The $\alpha$ parameter tunes homophily: a high $\alpha$ assigns labels based on proximity to centroids, while $\alpha=0$ assigns labels randomly. Negative $\alpha$ values also induce homophily, but nodes are labeled by the farthest centroid.

Fig.~\ref{fig:1} (a) illustrates the generation of graph-structured data with $N_n$ nodes, represented as blue circles, and $N_f$ features, depicted as purple rounded squares. The top part represents the unipartite network $\mathcal{G}_n$, connecting nodes generated with the $\mathbb{S}^1$ model. The positions of the cluster centroids are shown as $\theta_X$ and $\theta_Y$. The sketch is generated with a high value of $\alpha$ so that nodes are assigned labels of the nearest cluster centroids. Below $\mathcal{G}_n$, we depict the bipartite-$\mathbb{S}^1$ model where nodes and features share the same similarity space. With these two models, we can generate the synthetic graphs with the class labels and the binarized feature vectors shown at the bottom.

In Fig.~\ref{fig:1}(b) we depict the hyperbolic representation of one synthetic dataset generated with our model with realistic parameters. Typically, this corresponds to highly clustered, small-world, and heterogeneous degrees in the case of $\mathcal{G}_n$. As for $\mathcal{G}_{n,f}$, typical topologies have quite homogeneous nodes' degree distribution and very heterogeneous features degree distribution~\cite{aliakbarisani2023featureenriched}. Each column in Fig.~\ref{fig:1}(b) shows the two networks $\mathcal{G}_n$ and $\mathcal{G}_{n,f}$ with different values of the parameter $\alpha$. In the bipartite network, nodes with homogeneous degrees are closer to the edge of the disk, while features exhibiting heterogeneous degrees are closer to the center of the circle, connected to nodes within the same angular sector for higher value of ${|\alpha|}$.

\section{HypNF benchmarking framework} \label{sec:Benchmarking framework} 
The HypNF benchmarking framework depicted in Fig.~\ref{fig:1} combines the $\mathbb{S}^1/\mathbb{H}^2$ and bipartite-$\mathbb{S}^1/\mathbb{H}^2$ models within a unified similarity space. Additionally, it incorporates a 
method for label assignment. This integration facilitates the creation of networks exhibiting a wide range of structural properties and varying degrees of correlation between nodes and their features. Specifically, our framework allows us to control the following properties:
\begin{itemize}[leftmargin=*]
	\item \textbf{Degree distributions}. The framework allows us to generate networks with arbitrary node and feature degree distributions by fixing the sets of hidden degrees in $\mathcal{G}_n$ and $\mathcal{G}_{n,f}$. A useful choice is the power-law distribution 

\[
\rho(\kappa) =(\gamma-1) \kappa_0^{\gamma-1} \kappa^{-\gamma}\; \;  \mbox{with} \; \; \kappa_0=\frac{\gamma-2}{\gamma-1} \langle k \rangle
\]
and parameter $\gamma$ controlling the heterogeneity of the degree distribution. This leaves us with three parameters $\gamma$, $\gamma_n$, $\gamma_f$ fixing the hidden degree distributions in the $\mathbb{S}^1$ and bipartite-$\mathbb{S}^1$ models, respectively.
	
	\item \textbf{Network average connectivity}. Another critical factor is the network average connectivity. We have the freedom to choose the average degree of the network $\mathcal{G}_n$, $\langle k \rangle$, as well as the average degree of nodes in the bipartite node-feature network $\mathcal{G}_{n,f}$, $\langle k_n \rangle$, by changing $\mu$ and $\mu_b$ parameters in Eqs.~\eqref{eq:mu} and~\eqref{eq:mub}. Notice that the average degree of features is fixed by the identity $\langle k_f \rangle=\frac{N_n}{N_f}\langle k_n \rangle$.
	
	\item \textbf{Clustering and topology-features correlations}. The level of clustering is regulated by parameters $\beta$ and $\beta_b$, binding the topology of both $\mathcal{G}_n$ and $\mathcal{G}_{n,f}$ to the common similarity space	through the triangle inequality. A higher value of $\beta$ implies that pairs of nodes that are close in the hyperbolic space (with small $\chi$) are more likely to get connected in $\mathcal{G}_n$. Likewise, a high value $\beta_b$ indicates that nodes and features separated by a short hyperbolic distance have a high probability to be connected in $\mathcal{G}_{n,f}$. Consequently, when $\beta>1$ and $\beta_b>1$ similar nodes tend to be connected and, at the same time, share common features. Thus, $\beta$ and $\beta_b$ not only influence the clustering in the networks but also adjust the correlation between topology and node features. We should stress that the ability to directly control the level of clustering, and thereby topology-feature correlations, is one of the key distinctive features of HypNF as compared to other benchmarking schemes.
	
	\item \textbf{Homophily}. For the task of NC, $\mathcal{N}_l$ identifies the number of classes and $\alpha$ governs the strength of the nodes' labels concentration around their centroids, allowing us to control the level of homophily in the network, defined as~\cite{Pei2020Geom-GCN}
	
	\begin{equation}
		\label{eq:homophily}
		\mathcal{H} = \frac{1}{N_n} \sum_{i=1}^{N_n} \frac{n_l(i)}{k_i},
	\end{equation}
	where $n_l(i)$ is the number of neighbors of node $i$ with the same label as $i$ and $k_i$ its degree. A high value of $\mathcal{H}$ indicates a tendency for similar nodes to connect with each other. Fig.~\ref{fig:homophily_synthetic}(a) in Appendix~\ref{APX:Homophily} depicts the relation between $\mathcal{H}$ and the parameter $\alpha$, where  a monotonic dependence on $|\alpha|$ can be observed. However, this relation is not symmetric for $\mathcal{N}_l>2$. The lower $\mathcal{H}$ for negative $\alpha$ stems from the angular organization of the labeled nodes. Indeed, Fig.~\ref{fig:homophily_synthetic}(b) in Appendix~\ref{APX:Homophily} shows that for the high positive $\alpha$, the distance between two maximally distant nodes is lower compared to the same negative $\alpha$ value. Notice that $\alpha=0$ corresponds to a maximally random assignment of labels, which results in the expected homophily parameter taking the value $\langle \mathcal{H} \rangle=1/\mathcal{N}_l$.
\end{itemize}

Leveraging the HypNF model with varying parameters, our benchmarking framework generates diverse graph-structured data. This allows for the evaluation of graph machine learning models on networks with different connectivity patterns and correlations between topology and node features. For tasks like NC and LP, the framework facilitates fair model comparisons, helping to assess a novel GNN against state-of-the-art architectures and providing insights into the data's impact on performance.

\begin{figure*}[t!]
	\centering
	\includegraphics[width=0.8\linewidth]{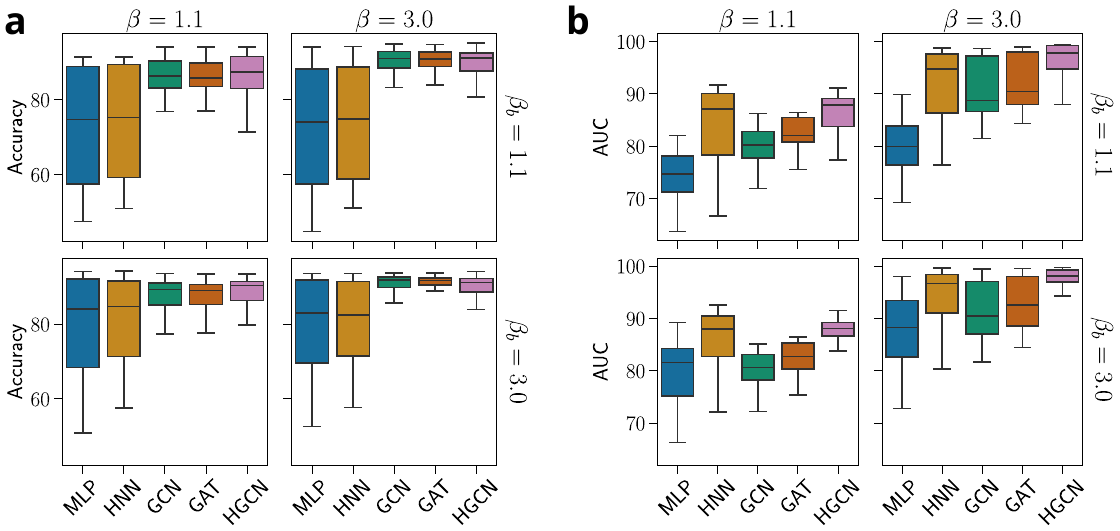}
	\caption{The impact of topology-feature correlation controlled by $\beta$ and $\beta_b$ on the performance of graph machine learning models in two tasks: (a) node classification and (b) link prediction. We set $\mathcal{N}_L=6$ and $\alpha=10$ for the NC task. The box ranges from the first quartile to the third quartile. A horizontal line goes through the box at the median. The whiskers go from each quartile to the minimum or maximum. The results are averaged over all other parameters.}
	\label{fig:2}
\end{figure*}

\section{Experiments} \label{sec: Experiments}

\subsection{Parameter Space}
We use the HypNF benchmarking framework introduced in Section~\ref{sec:Benchmarking framework} and customize its parameters to generate a collection of datasets with $N_n = 5000$ nodes and $N_f=2000$ features, displaying a wide spectrum of properties. The degree distributions for nodes in $\mathcal{G}_n$ and for nodes and features in $\mathcal{G}_{n,f}$ are tailored to exhibit either heterogeneity or homogeneity. This is achieved by setting $\gamma, \gamma_n,\gamma_f$ to the values $\{2.1, 3.5\}$. The average degrees vary between low and high values by adjusting $\langle k \rangle$, and $\langle k_n \rangle$ to $\{3, 30\}$. The clustering coefficient in both networks, along with the degree of correlation between them, is set to range between low and high values by fixing $\beta$ and $\beta_b$ to $\{1.1, 3\}$. For the task of NC, the number of labels are $\mathcal{N}_l = \{2, 3, 6, 10\}$. Homophily of node labels is varied by setting $\alpha = \{-1, 1, 5, 10\}$.

For each unique combination of parameter values, we generate 10 network realizations with node features, resulting in a comprehensive benchmark for model evaluation. Consequently, for the LP task, machine learning models are tested on a benchmark comprising $2 ^ 7 \times 10 = 1280$ instances. In the case of NC, where we introduce two additional parameters, $\alpha$ and $\mathcal{N}_l$, the size of the benchmark is $2 ^ 7 \times 4^2 \times 10 = 20480$. In addition, we evaluate the performance of GNNs using AUC for the LP task~\cite{Melo2013}, and accuracy defined as the fraction of correctly classified nodes for NC task~\cite{bishop2006pattern}.

\subsection{Machine learning models}
In this work, we focus on two primary methodologies: feature-based methods, which entail node embedding based on their features, and GNNs, which integrate both features and network topology. 

\begin{itemize}[leftmargin=*]
	\item \textbf{MLP}: A vanilla neural network transforms node feature vectors through linear layers and non-linear activations to learn embeddings in Euclidean space.
	\item \textbf{HNN}~\cite{ganea2018hyperbolic}: A variant of MLP that operates in hyperbolic space to capture complex patterns and hierarchical structures.
	\item \textbf{GCN}~\cite{Kipf2017Semi}: A pioneering model that averages the states of neighboring nodes at each iteration.
	\item \textbf{GAT}~\cite{velickovic2018graph}: A model that uses attention mechanisms to assign different importances to different nodes in a neighborhood.
	\item  \textbf{HGCN}~\cite{Chami2019Hyperbolic}: A model that integrates hyperbolic geometry with graph convolutional networks to capture complex structures in graph data more effectively.
\end{itemize}

Table~\ref{tab:hyperparameters} in Appendix \ref{APX:Hyperparameters} lists the hyperparameters for training. In the LP task, links are split into training (85\%), validation (5\%), and test (10\%) sets. For the NC task, nodes are distributed as 70\% training, 15\% validation, and 15\% test~\cite{Chami2019Hyperbolic}. Both tasks follow the methodology in~\cite{Chami2019Hyperbolic}, with results averaged over five test-train splits. Models were trained on an NVIDIA GeForce RTX 3080 GPU using Python 3.9, CUDA 11.7, and PyTorch 1.13.

\vspace{-0.35cm}
\section{Results}\label{Sec:results}

We begin by analyzing how topology-feature correlation affects the performance of machine learning models by varying parameters $\beta$ and $\beta_b$, which control the coupling between $\mathcal{G}_n$, $\mathcal{G}_{n,f}$, and the shared metric similarity space. For both NC and LP tasks, Fig.~\ref{fig:2} shows that higher topology-feature correlation ($\beta = \beta_b = 3$) improves performance for both feature-based and GNN models compared to lower correlation ($\beta = \beta_b = 1.1$). In the NC task, feature-based models do not respond to changes in $\beta$ and perform better with higher $\beta_b$. Conversely, GNN models, utilizing information from both $\mathcal{G}_n$ and $\mathcal{G}_{n,f}$, benefit from high $\beta$ and $\beta_b$ values, not only in terms of average performance but also in terms of the spread around this average.

We measured the performance of machine learning models by varying the clustering level in $\mathcal{G}_n$ (adjusted by $\beta$) and the average number of features per node in the bipartite network, $\langle k_n \rangle$. For the NC task, Fig.~\ref{fig:beta-kmean-n}(a) in Appendix~\ref{APX:Performance analysis} shows significant performance differences between feature-based and GNN models. In bipartite networks with low $\langle k_n \rangle$, GNN models outperform feature-based ones. As $\langle k_n \rangle$ increases (bottom row), all models' accuracy improves, converging to almost similar levels. Here, the simplest model, MLP, performs nearly as well as the most sophisticated, HGCN. Note that these results fix $\beta$ and $\langle k_n \rangle$, averaging over other parameters. Fig.~\ref{fig:beta-kmean-n}(b) in Appendix~\ref{APX:Performance analysis} compares these models in terms of AUC for the LP task. The models show a significant shift in relative performance with varying $\langle k_n \rangle$. For high $\langle k_n \rangle$, models in hyperbolic space (HNN and HGCN) perform better. Notably, the feature-based model, HNN, matches or exceeds the performance of HGCN, especially at low $\beta$.

\begin{figure*}[t!]
	\centering
	\includegraphics[width=\linewidth]{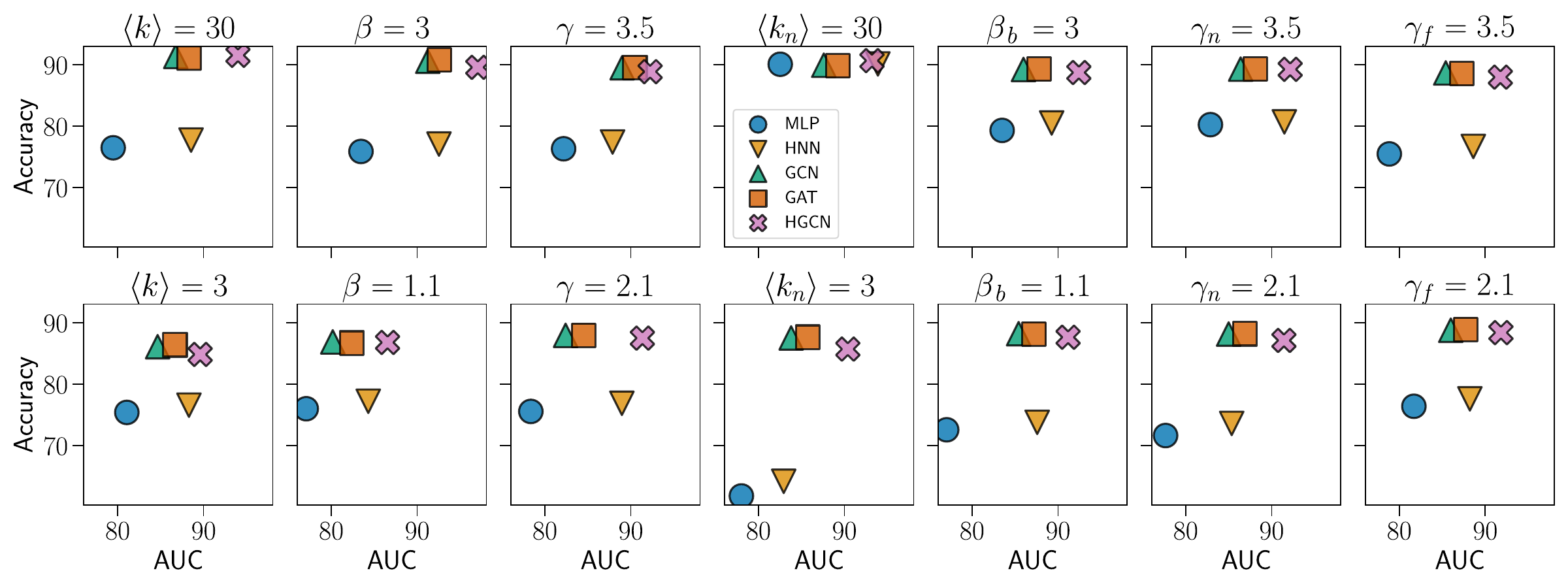}
	\caption{Impact of each individual parameter on the performance of NC  and LP. In the case of NC, we set $\mathcal{N}_L=6$ and $\alpha=10$.} 
	\label{fig:3}
\end{figure*}
In Fig.~\ref{fig:3}, we present a summary of model performance on both downstream tasks, varying parameters individually between high and low values while averaging over others. Across most cases, HGCN outperforms or matches others in the LP task, and achieves competitive accuracy in the NC task, akin to other graph-based methods. In addition, for the NC task, the feature-based methods (MLP and HNN) are barely sensible to $\mathbb{S}^1$ model parameters, i.e., $\langle k \rangle$, $\beta$ and $\gamma$. However, for the LP task, the topology-features correlation strongly impacts their AUC values, leading to MLP and HNN being sensitive to the parameters of the bipartite network, including $\beta_b$, $\gamma_n$, and $\gamma_f$, with a specific emphasis on $\langle k_{n} \rangle$. In contrast, GNNs are particularly responsive to variations in $\langle k_{n} \rangle$. It is worth mentioning that these observations highlight the sensitivity of graph machine learning methods to individual parameters. However, in real-world scenarios, the collective interplay of all parameters influences the overall performance of the models, see the detailed analysis in Appendix~\ref{APX:Performance analysis}.

Another important factor that is usually overlooked concerns the fluctuations of the performance of machine learning models, which provides insights into their robustness and reliability. We address this problem by analyzing the difference in the standard deviation of the accuracy and AUC for a given set of parameters. In Fig.~\ref{fig:Fluctuation_ML_models} of Appendix~\ref{APX:Fluctuation}, we show how the fluctuation changes when each parameter is set to high or low. Let us take HGCN as an example. For the high average degree $\left<k\right>$, the accuracy and AUC display lower standard deviations as compared to low average degree, which means that HGCN is more robust to other parameters when the average degree is high. Moreover, a homogeneous degree distributions ($\gamma=3.5$) results in a broader spread of AUC but not of accuracy. GCN and GAT display similar fluctuation behaviors, whereas for the feature-based models (MLP, HNN), the bipartite-$\mathbb{S}^1$ parameters dictate their sensibility to other parameters.

\begin{figure}[t]
	\centering
	\includegraphics[width=0.65\linewidth]{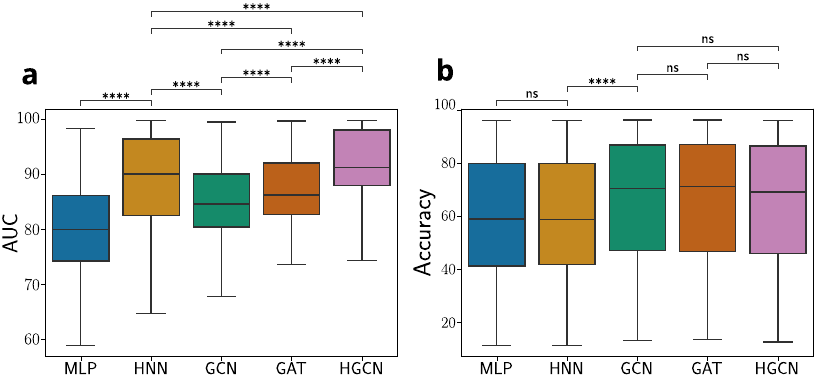}	
	\caption{Aggregated results for (a) link prediction and (b) node classification tasks. The annotations denote the statistical significance levels derived from the Mann-Whitney U test, a non-parametric test suitable for comparing two independent samples, especially when the data distribution is not assumed to be normal. This test evaluates if there is a statistically significant difference in the median performance scores between two different models.
	The label $\textit{ns}$ signifies p-value $\le 1$, $\textbf{****}$ corresponds to p-value $\le 10^{-4}$.}
	\label{fig:4}
\end{figure}

The global average picture of the performance of the analyzed models is shown in Fig.~\ref{fig:4}, where we averaged results over all the parameters in the HypNF benchmarking framework. This analysis sheds light on model selection when no structural information is available about the data. We observe the superiority of the hyperbolic-based models in capturing essential network properties and connectivities for the LP task (Fig.~\ref{fig:4}(a)). Yet, HGCN outperforms HNN with statistical significance. As for the NC task, results are more ambiguous, with GNNs outperforming feature-based models overall. However, the distinctions in performance within the GNNs are not substantial. Hence, when dealing with unseen data, the simplicity of the GCN model makes it a more efficient choice.

Finally, in Appendix~\ref{APX:Performance analysis}, we carry out a comprehensive analysis of the performance of the selected models, focusing on all parameters within the $\mathbb{S}^1/\mathbb{H}^2$ (Figs.~\ref{fig:All_param_nc}(a) and~\ref{fig:All_param_lp}(a)), bipartite-$\mathbb{S}^1/\mathbb{H}^2$ (Figs.~\ref{fig:All_param_nc}(b) and~\ref{fig:All_param_lp}(b)), and the combination of the two (Figs.~\ref{fig:mixed_param_nc} and~\ref{fig:mixed_param_lp}). Moreover, Fig.~\ref{fig:nc_nlabels_alpha} reveals the dependence of the number of labels and the homophily level in terms of accuracy.
\vspace{-0.6cm}

\section{Conclusion} \label{sec:Conclusion}
In this study, we introduced HypNF benchmarking framework for graph machine learning, and in particular for graph neural networks. This framework, leveraging the $\mathbb{S}^1/\mathbb{H}^2$ and bipartite-$\mathbb{S}^1/\mathbb{H}^2$ models, enables the generation of synthetic networks with controllable properties such as degree distributions, average degrees, clustering coefficients, and homophily levels. Our findings underscore the significance of topology-feature correlations in influencing GNN performance. Specifically, models embedded in the hyperbolic space, with its intrinsic hierarchical structure, outperformed the rest of the models in the LP task. This research contributes to the broader understanding of graph machine learning, providing insights into the suitability of various models under different network conditions. Furthermore, our benchmarking framework serves as a valuable tool for the community, enabling fair and standardized comparisons of new GNN architectures against established models.
A promising direction for future work is to extend the bipartite-$\mathbb{S}^1/\mathbb{H}^2$ model to incorporate weighted features. This will enable HypNF to generate networks with non-binary features, allowing for more nuanced and flexible representations~\cite{Allard2017}. Additionally, the framework can be extended to accommodate community structures by using a non-homogeneous distribution of nodes within the $\mathbb{S}^1/\mathbb{H}^2$ similarity space, as done in~\cite{Zuev2015} and~\cite{Garcia-Perez2018}.

\begin{ack}
We acknowledge support from: Grant TED2021-129791B-I00 funded by MCIN/AEI/10.13039/501100011033 and the ``European Union NextGenerationEU/PRTR''; Grant PID2022-137505NB-C22 funded by MCIN/AEI/10.13039/501100011033; Generalitat de Catalunya grant number 2021SGR00856. R.~J. acknowledge support from the fellowship FI-SDUR funded by Generalitat de Catalunya. M. B. acknowledges the ICREA Academia award, funded by the Generalitat de Catalunya.
\end{ack}

\bibliographystyle{ACM-Reference-Format}
\bibliography{refs}
\clearpage

\newpage
\appendix

\section{Empirical validation of HypNF} \label{APX:Empirical validation}
\begin{figure*}[h]
	\centering
	\includegraphics[width=\linewidth]{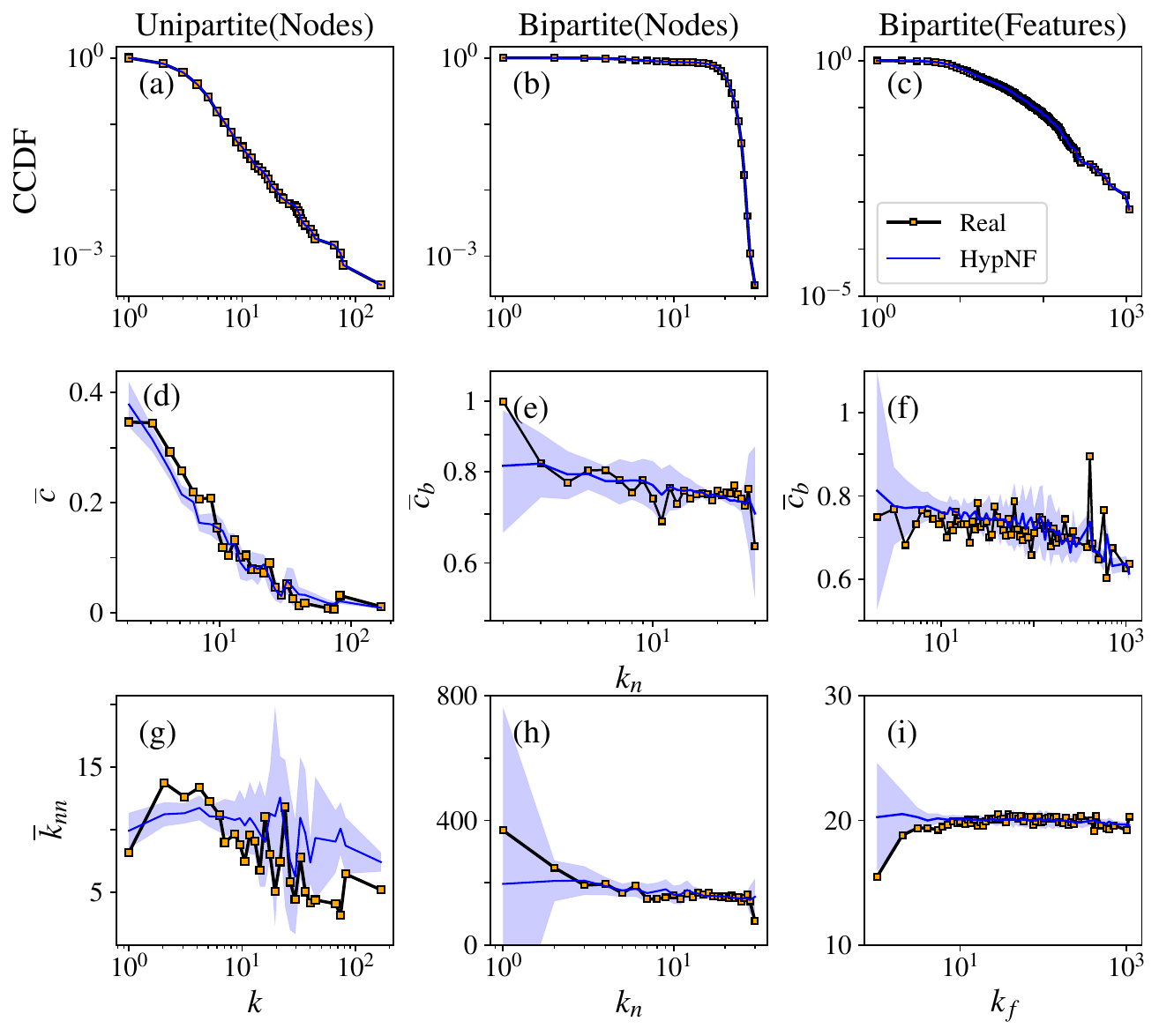}	
	\caption{Topological properties of Cora dataset and its synthetic counterpart generated by HypNF. (a)-(c) Complementary cumulative degree distribution of nodes in $\mathcal{G}_n$  and those of nodes and features in $\mathcal{G}_{n,f}$. (d)-(f) clustering spectrum of nodes as a function of node degrees in $\mathcal{G}_n$ and bipartite clustering sepctrum of nodes and features as a function of nodes and features degrees in $\mathcal{G}_{n,f}$. (g)-(i) average nearest neighbors degree function in $\mathcal{G}_n$ and $\mathcal{G}_{n,f}$. The clustering coefficient, $c$, for each node in $\mathcal{G}_n$ is calculated as the fraction of connected pairs among its neighboring nodes relative to the total possible pairs of neighbors. To compute the bipartite clustering, $c_b$, for a given node, each pair of its neighboring features is considered connected if they share at least one common node, apart from the node in question. The clustering coefficient is then computed similarly to the clustering coefficient in a unipartite network. This definition also applies to the bipartite clustering coefficient of features. Finally, $k_{nn}$ in both $\mathcal{G}_n$ and $\mathcal{G}_{n,f}$ quantifies the average degree of the neighbors for a given node or feature. Exponential binning is used in the computation $\overline{c}$, $\overline{c}_b$ and $\overline{k}_{nn}$. The blue shaded region indicates the two-$\sigma$ intervals around the mean, derived from 100 realizations of HypNF.}
	\label{fig:Cora_Properties}
\end{figure*}

\newpage

\section{Degree distribution and clustering control in HypNF} \label{APX:Degree&Clustering control}
\begin{figure*}[h]
	\centering
	\includegraphics[width=\linewidth]{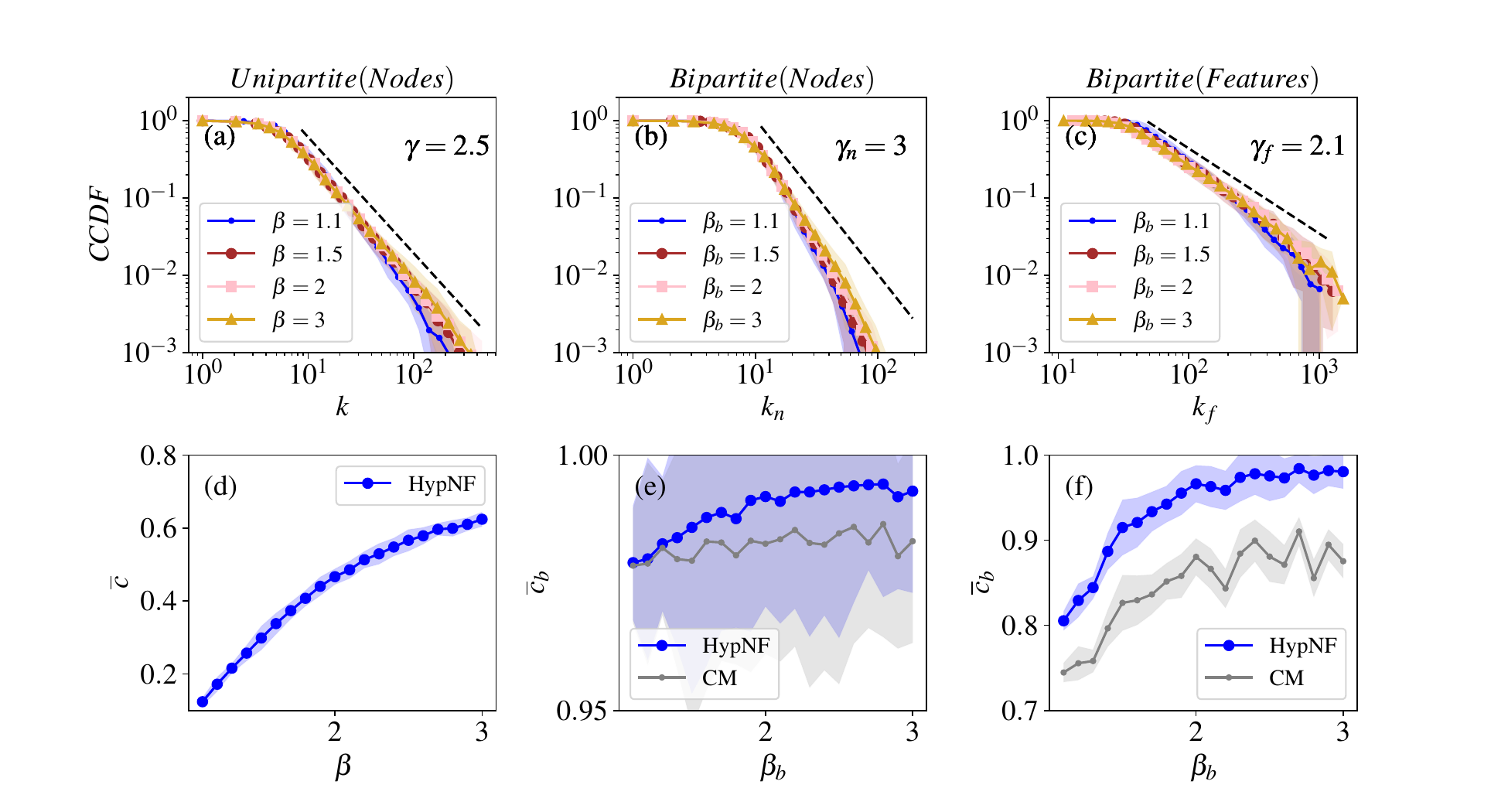}	
	\caption{Independent control of degree distribution and clustering coefficient in synthetic networks generated by HypNF. (a)-(c) Complementary cumulative degree distribution of nodes in $\mathcal{G}_n$ ($N = 2000$, $\gamma = 2.5$, and $\langle k \rangle = 10$) and those of nodes and features in $\mathcal{G}_{n,f}$ ($N_n = 2000$, $N_f = 200$, $\langle k_n \rangle =10$, $\gamma_n = 3$, and $\gamma_f=2.1$) for different values of $\beta$ and $\beta_b$. The dashed black lines are guides for the eyes corresponding to the power-law exponents. (d)-(e) Clustering and bipartite clustering coefficients as a function of $\beta$ and $\beta_b$. The gray curves in (e) and (f) illustrate the bipartite clustering in the randomized versions of the bipartite synthetic networks, generated according to the configuration model (CM).}
	\label{fig:Param_independency}
\end{figure*}

\newpage
\section{Hyperparameters of the machine learning models} \label{APX:Hyperparameters}

\begin{table*}[h]
	\caption{Hyperparameters of the machine learning models.}
	\label{tab:hyperparameters}
	\centering
	\begin{tabular}{ccccccc}
		\toprule
		Model & Dropout & Weight Decay & Optimizer & Activation & Layers  & Dimensions \\ \midrule
		MLP                & 0.2 & 0.001   & Adam &  - &  2  & 16 \\
		HNN                & 0.2 & 0.001   & Adam &  - &  2  & 16 \\
		GCN                & 0.2 & 0.0005  & Adam &  ReLU &  2  & 16 \\
		GAT                & 0.2 & 0.0005  & Adam &  ReLU &  2  & 16 \\
		HGCN               & 0.2 & 0.0005  & Adam &  ReLU &  2 & 16 \\
		\bottomrule
	\end{tabular}%
\end{table*}

\section{Fluctuations in the performance of machine learning models} \label{APX:Fluctuation}

\begin{figure}[h]
	\centering
	\includegraphics[width=0.85\linewidth]{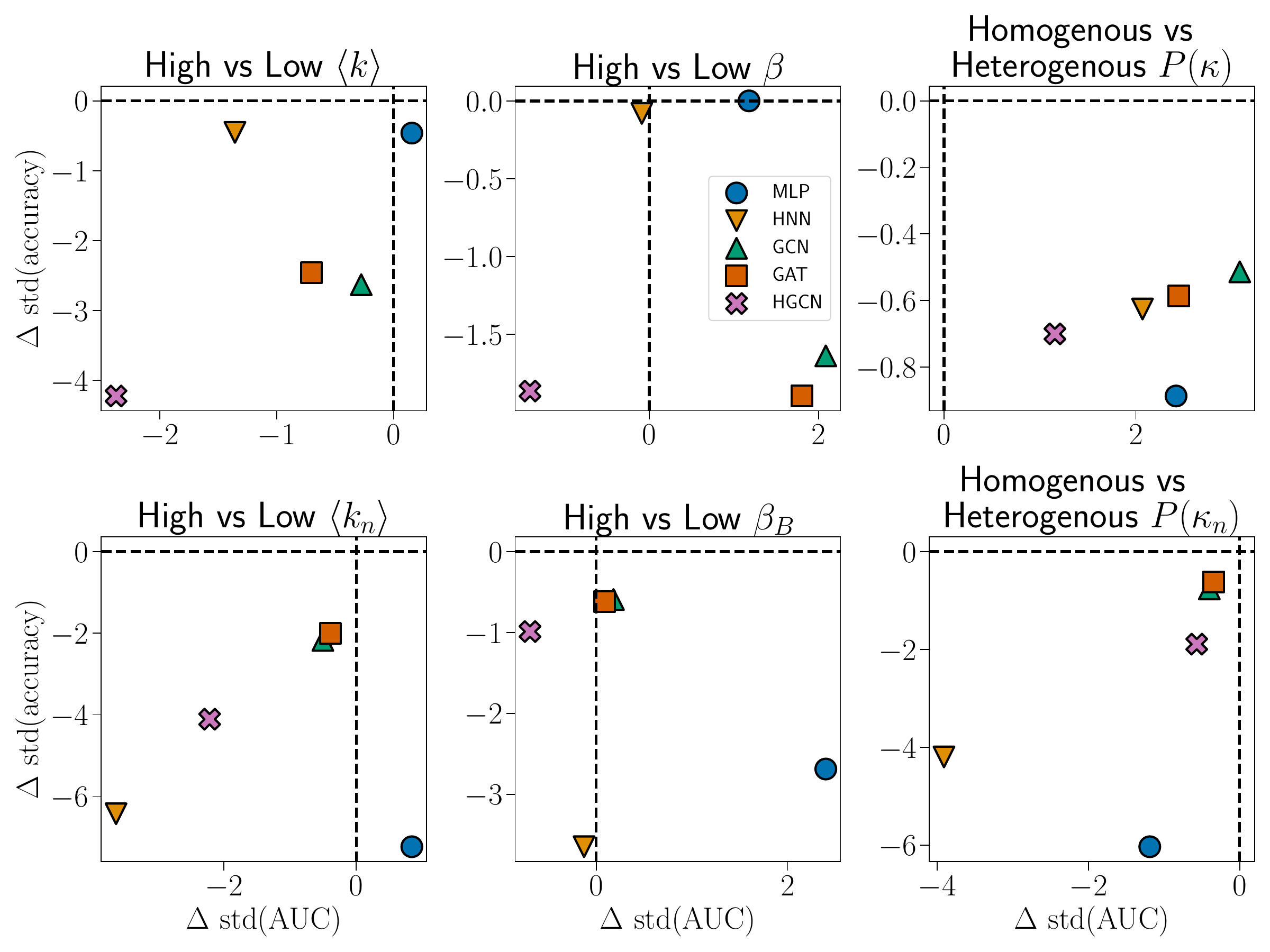}	
	\caption{Fluctuations of performance of the machine learning models. The difference of the standard deviation between high and low value of each parameter. We set $\mathcal{N}_L=6$ and $\alpha=10$ for the NC task. The results are averaged over all other parameters.}
	\label{fig:Fluctuation_ML_models}
\end{figure}

\newpage
\section{Homophily in the synthetic networks} \label{APX:Homophily}

\begin{figure*}[h]
	\centering
	\includegraphics[width=\linewidth]{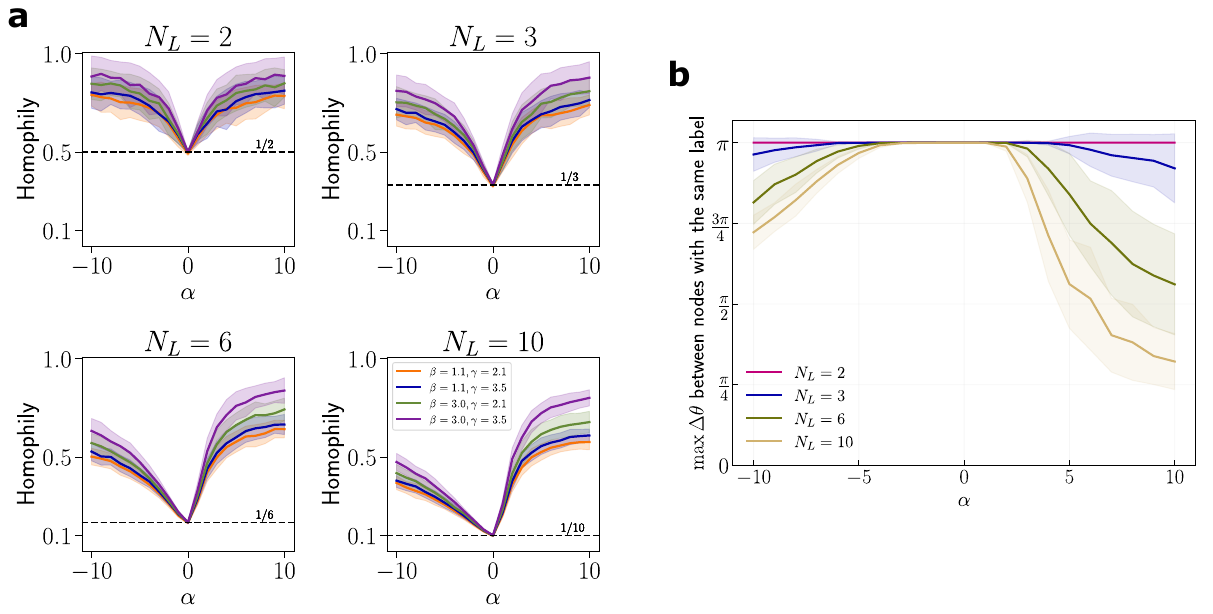}
	\caption{(a) Value of homophily in the synthetic networks. Parameters of the networks: $N=1000, \left<k\right>=30$. The black horizontal line indicates the random case, i.e., for $\alpha=0$ which based on Eq.~\ref{eq:1} gives us $\mathcal{H}=1/N_L$ where $N_L$ is the number of labels. The results are averaged over 100 realizations. (b) Maximum angular distance between nodes in the same community in function of parameter $\alpha$. The results are averaged over 100 realizations.}
	\label{fig:homophily_synthetic}
\end{figure*}


\section{Exploring the parameters' space} \label{APX:Performance analysis}
Here, we explore the impact of parameters within the HypNF benchmarking framework including the parameters in $\mathbb{S}^1/\mathbb{H}^2$ and bipartite-$\mathbb{S}^1/\mathbb{H}^2$ models, as well as the combinations of both on the performance of graph machine learning techniques.

\begin{figure*}[h]
	\centering
	\includegraphics[width=\linewidth]{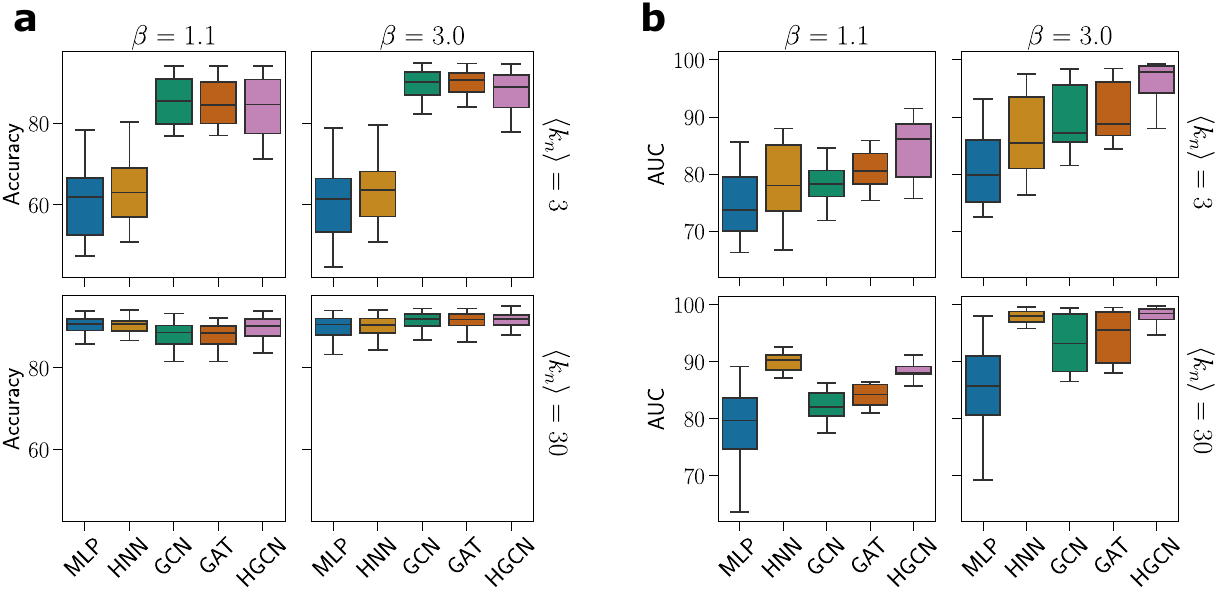}
	\caption{The impact of the level of clustering $\beta$ and average degree of nodes in the bipartite network $\langle k_n \rangle$ on the performance of machine learning models in two tasks: (a) node classification and (b) link prediction. We set $\mathcal{N}_L=6$ and $\alpha=10$ for the NC task. The box ranges from the first quartile to the third quartile. A horizontal line goes through the box at the median. The whiskers go from each quartile to the minimum or maximum. The results are averaged over all other parameters.}
	\label{fig:beta-kmean-n}
\end{figure*}

\begin{figure*}[h]
	\centering
	\includegraphics[width=\linewidth]{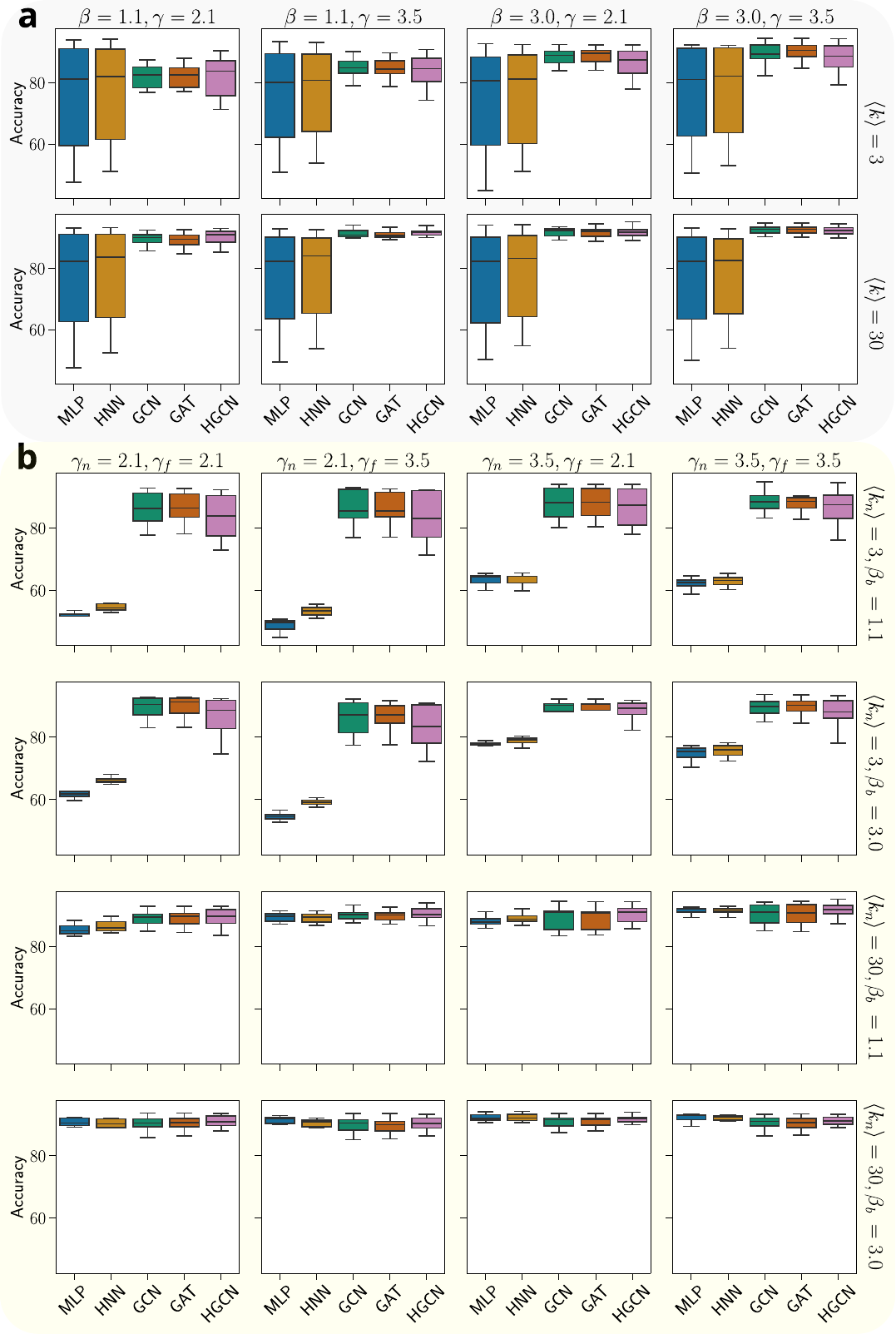}	
	\caption{Performance comparison of machine learning models on NC task with respect to (a) $\mathbb{S}^1/\mathbb{H}^2$ model parameters and (b) bipartite-$\mathbb{S}^1/\mathbb{H}^2$ model parameters. We set $\mathcal{N}_L=6$ and $\alpha=10$.}
	\label{fig:All_param_nc}
\end{figure*}

\begin{figure*}[h]
	\centering
	\includegraphics[width=\linewidth]{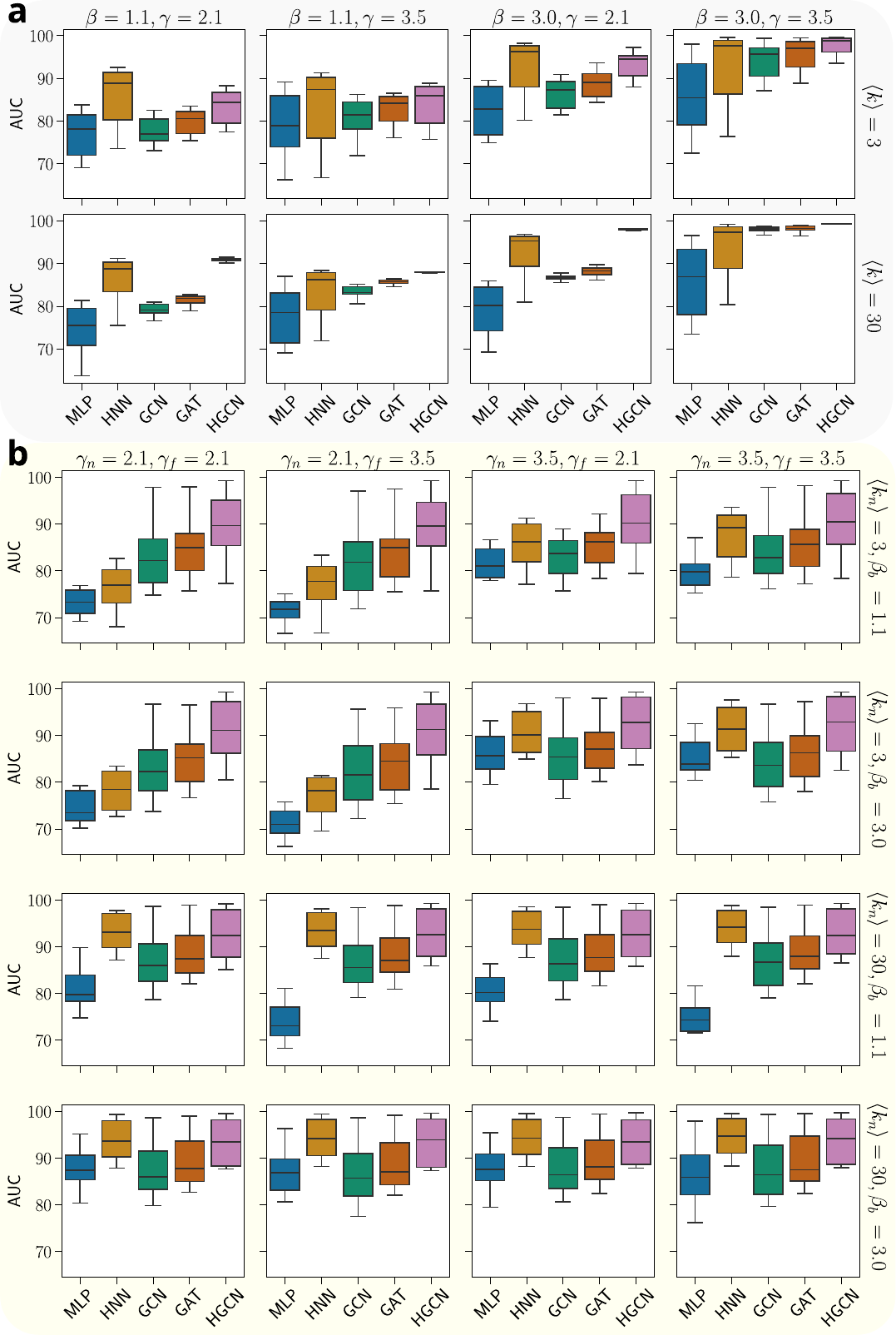}	
	\caption{Performance comparison of machine learning models on LP task with respect to (a) $\mathbb{S}^1/\mathbb{H}^2$ model parameters and (b) bipartite-$\mathbb{S}^1/\mathbb{H}^2$ model parameters.}
	\label{fig:All_param_lp}
\end{figure*}

\begin{figure*}[h]
	\centering
	\includegraphics[width=\linewidth]{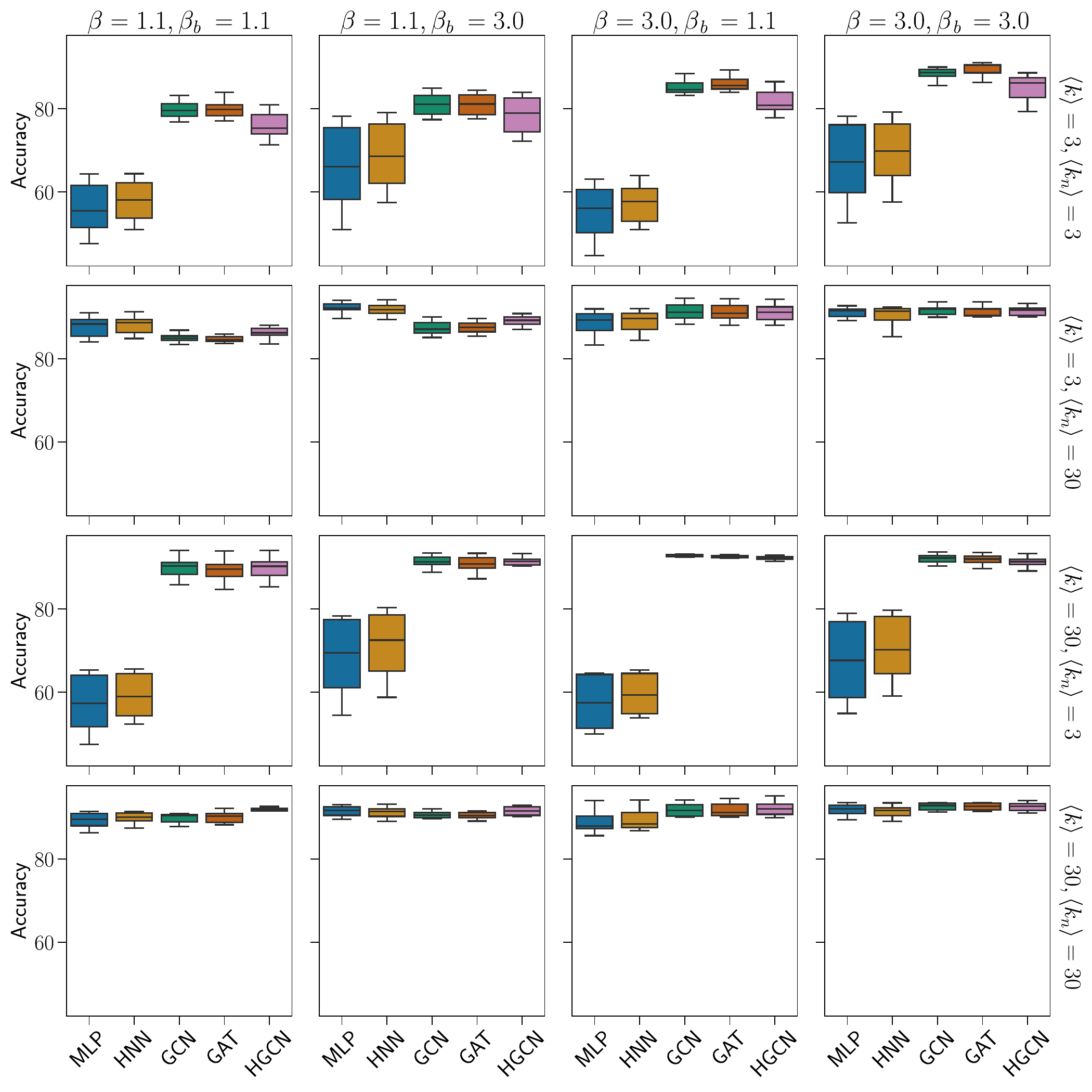}	
	\caption{Performance comparison of machine learning models on NC task with respect to a combination of parameters in the $\mathbb{S}^1/\mathbb{H}^2$ and bipartite-$\mathbb{S}^1/\mathbb{H}^2$ models. We set $\mathcal{N}_L=6$ and $\alpha=10$.}
	\label{fig:mixed_param_nc}
\end{figure*}

\begin{figure*}[h]
	\centering
	\includegraphics[width=\linewidth]{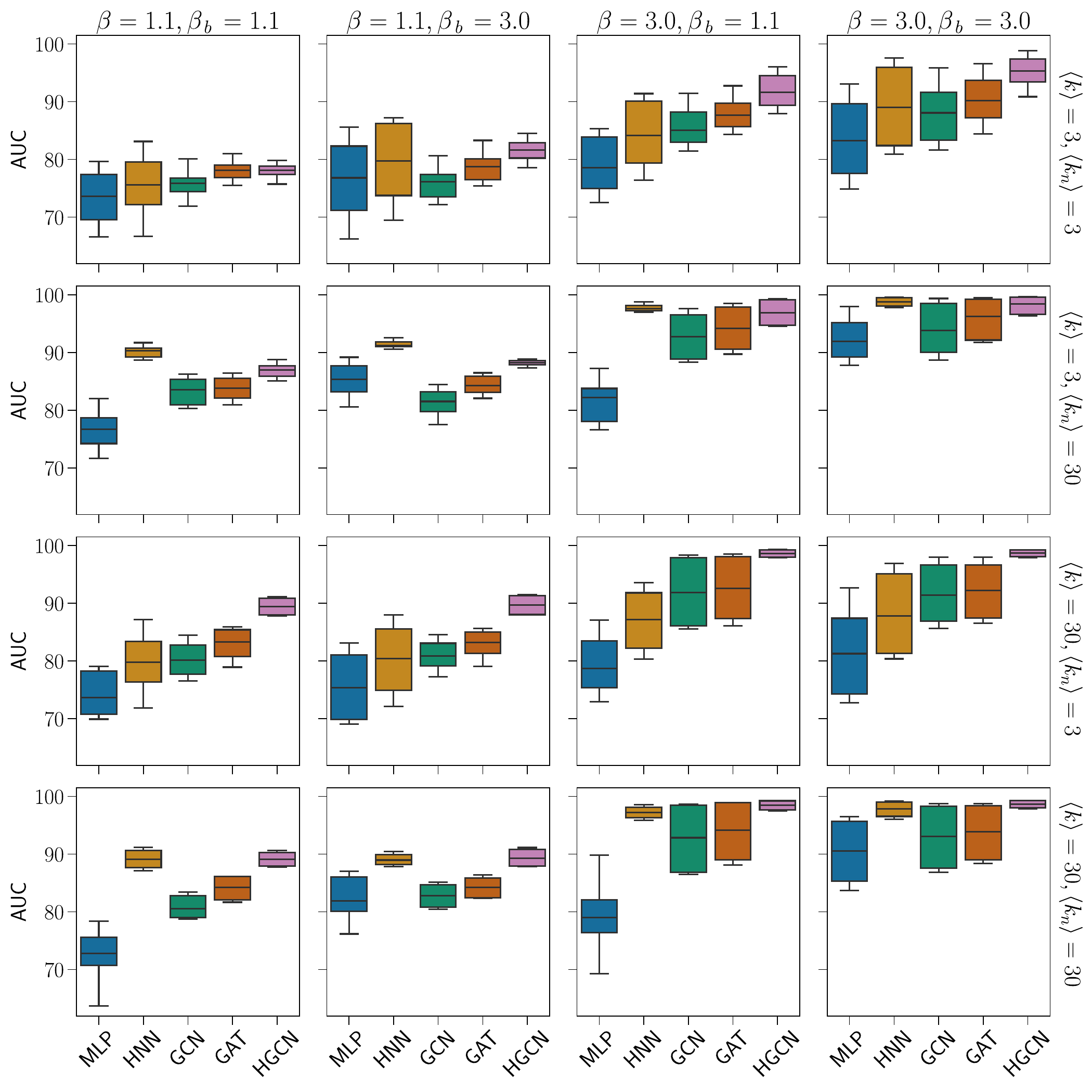}	
	\caption{Performance comparison of machine learning models on LP task with respect to a combination of parameters in the $\mathbb{S}^1/\mathbb{H}^2$ and bipartite-$\mathbb{S}^1/\mathbb{H}^2$ models.}
	\label{fig:mixed_param_lp}
\end{figure*}

\begin{figure*}[h]
	\centering
	\includegraphics[width=\linewidth]{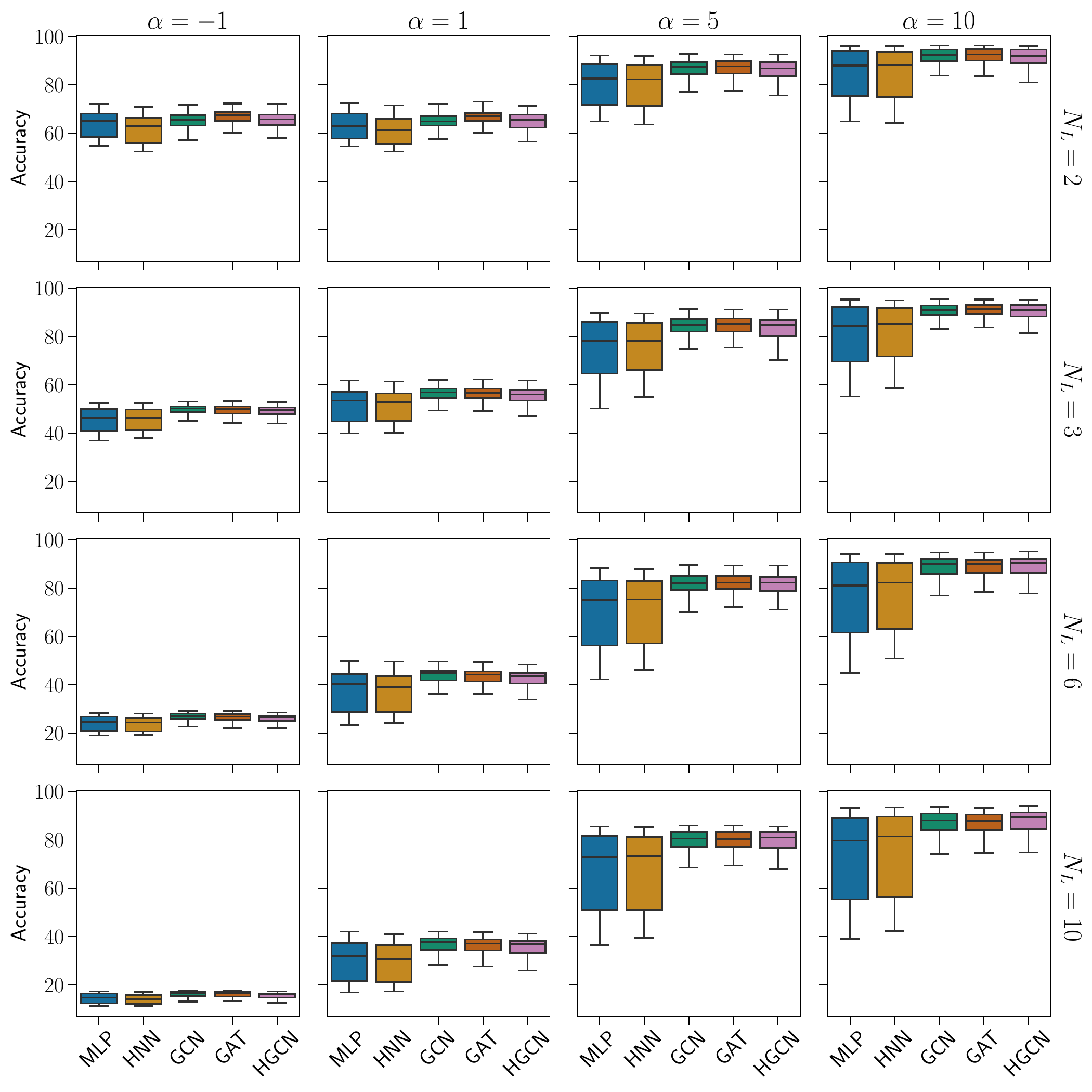}	
	\caption{Performance comparison of machine learning models on the NC task with respect to a combination of $\mathcal{N}_l$ and $\alpha$ parameters averaged over all other parameters.}
	\label{fig:nc_nlabels_alpha}
\end{figure*}

\end{document}